%% file: main.tex
\documentclass[cameraready]{Interspeech}

\title{Fair Cognitive Impairment Detection Through Unlearning}

\author[]{William}{Nguyen}
\author[orcid=0009-0006-3376-6549]{Jiali}{Cheng}
\author[orcid=0000-0002-3485-6697]{Hadi}{Amiri}

\address{
    University of Massachusetts Lowell, USA
}

\email{\{htnguyen, jcheng2, hadi\}@cs.uml.edu}


\usepackage{comment}
\usepackage{xspace}
\usepackage{todonotes}
\usepackage{multirow}

\newcommand{\method}{FMD\xspace}

\begin{document}

\maketitle

\input{000abstract}
\input{010intro}
\input{020related}

\input{030method}

\input{040experiments}

\input{050results}

\input{060conclusion}


\bibliographystyle{IEEEtran}
\bibliography{reference,anthology}

\end{document}

%% file: 000abstract.tex
\begin{abstract}
    Mild Cognitive Impairment (MCI) is a medical condition characterized by a noticeable decline in memory, language, or thinking abilities. MCI detection from spontaneous speech is promising for scalable screening. However, learned models often exploit demographic cues correlated with labels, resulting in a large performance gap across subgroups. We present a multimodal framework that combines
(i) cross-model fusion between modalities (speech, text, and image), and
(ii) unlearning using gradient reversal that discourages the shared embedding from encoding task-irrelevant demographic attributes. Evaluated on the multilingual benchmarks TAUKADIAL and PREPARE, our method outperforms the state-of-the-art multilingual and multimodal baseline in MCI classification while substantially reducing the performance gap across patient subgroups (sex and language). We further analyze transfer across datasets, showing that demographic unlearning helps learn more robust representations for MCI detection.
\href{https://github.com/CLU-UML/Fair-MCI-Detection}{Our code is here}.
\end{abstract}

%% file: 010intro.tex
\section{Introduction}

Speech-based assessment is a promising approach for screening cognitive impairment because spontaneous speech reflects cognitive and linguistic changes in lexical choice and diversity, syntactic complexity, disfluencies, and prosody~\cite{de2020artificial,luz21_interspeech}. 
However, real-world clinical speech datasets are typically small, heterogeneous, and demographically imbalanced, and machine learning models trained on them can learn spurious demographic correlations rather than true cognitive markers. Specifically, dementia speech benchmarks are known to be vulnerable to confounding by demographic factors such as sex~\cite{cheng24c_interspeech,azadmaleki2025speechcare,gulzar2026biasfairnessselfsupervisedacoustic}. As a result, the same model may perform well for one demographic subgroup but degrade substantially for another, raising concerns about reliability and equitable deployment.

Existing approaches for bias mitigation can identify both known~\cite{clark-etal-2019-dont,karimi-mahabadi-etal-2020-end,modarressi-etal-2023-guide} and previously unidentified~\cite{utama-etal-2020-towards,sanh2020learning} dataset biases. They mitigate biases through re-weighting examples~\cite{sanh2020learning,karimi-mahabadi-etal-2020-end}, learning robust representations~\cite{gao-etal-2022-kernel,du-etal-2023-towards}, identifying robust feature interaction patterns~\cite{wang-etal-2023-robust}, reducing the influence of biased model components~\cite{meissner-etal-2022-debiasing}, and applying data perturbation~\cite{cheng-amiri-2024-fairflow}. 
Despite several recent works~\cite{cheng24c_interspeech,gulzar2026biasfairnessselfsupervisedacoustic}, fairness and bias mitigation is underexplored in speech-based cognitive assessment, particularly in multilingual and multimodal settings where demographic cues can appear in both acoustics (e.g., pitch, formants, speaking rate) and text (e.g., vocabulary, grammar, code-switching). A central challenge is therefore to learn representations that capture cognitive impairment signals while being invariant to protected demographic variables.

We address this problem by proposing \method, a fair MCI detection framework built on two core ideas. First, instead of relying on late concatenation of modalities, we use cross-attention fusion to enable multimodal representations to interact and align; this allows the model to emphasize complementary cognitive cues that may be expressed unevenly across modalities. Second, we incorporate an unlearning approach.
An auxiliary demographic classifier attempts to predict protected attributes from the shared embedding, while the unlearning module guides the encoders to remove demographic information that could act as a shortcut for the main diagnosis task (MCI status). This prioritizes task-predictive features over uninformative or irrelevant signals when learning representations. Our goal is not to erase clinically meaningful variations, but to reduce reliance on demographic identity as a proxy for MCI detection.\looseness-1
 
We evaluate \method on two multilingual benchmarks (TAUKADIAL~\cite{taukadial} and PREPARE~\cite{prepare}) and assess (i) overall MCI detection performance, (ii) performance gaps across sex and language subgroups, and (iii) robustness under distribution shift, quantified through model transferability across datasets. Experimental results show that \method improves average F1 on both datasets while reducing subgroup disparities, with larger improvements on TAUKADIAL (smaller, three modalities) compared to PREPARE (larger, two modalities).

The contribution of this paper is \method: a fair MCI detection model that fuses speech, text, and image modalities through cross-attention, with a gradient reversal component that unlearns demographic biases in the shared representations.
\vspace{-10pt}

%% file: 020related.tex
\section{Related Work}

\textbf{MCI detection}: Existing work has developed models to detect MCI and Alzheimer's disease from speech signals using semantic features~\cite{heitz-etal-2025-linguistic}, linguistic features~\cite{gkoumas-etal-2023-digital,gkoumas-etal-2023-reformulating}, training data augmentation~\cite{li-etal-2023-two,duan-etal-2023-cda}, and prompt learning~\cite{farzana-parde-2024-domain}. \cite{treder2024introduction} provides a review of LLM applications in dementia care, analyzes survey results from individuals with dementia and their supporters, and outlines priorities for AI-driven healthcare solutions.

\begin{figure*}[t]
    \centering
    \vspace{-20pt}
    \includegraphics[width=0.98\linewidth]{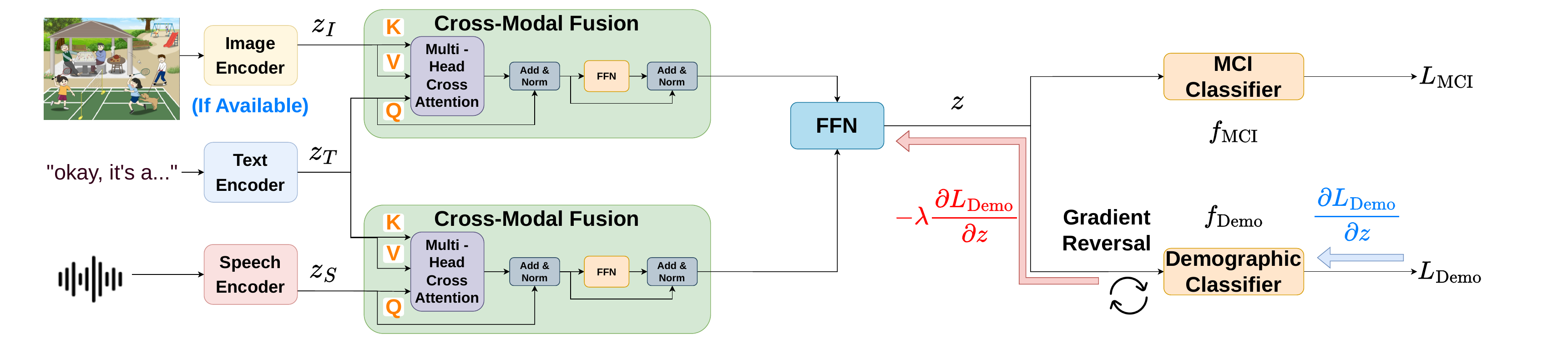}
    \caption{Architecture of \method.  
    1) The cross-modal (CM) fusion module followed by a feed-forward network (FFN) supports richer and finer-grained modality interaction and fusion compared to standard concatenation.
    2) The unlearning (UL) component removes task-irrelevant biases from the model, yielding fairer and more robust performance. This is achieved using an auxiliary demographic classifier $f_{\text{Demo}}$ that identifies spurious demographic features; the gradient of $f_{\text{Demo}}$ is reversed to unlearn these spurious features.
    }
    \label{fig:fig1}
\end{figure*}

\textbf{Bias in speech recognition}: 
Previous works~\cite{lin2024social,lin2024emo,pelloin2024automatic,lai2023exploring} highlight the importance of evaluating biases (sex, age, accent, and other demographics) that influence speech recognition models. \cite{koudounas2024contrastive} proposes a contrastive learning method to mitigate bias in different under-performing subgroups. \cite{zhang2022mitigating} proposes data augmentation and domain adversarial training methods to mitigate bias against non-native accents. \cite{kim2024automatic} proposes a multi-head model technique and data augmentation to mitigate speaker bias for children speech sound disorder detection. \cite{goria2024revealing} demonstrates the effect of confounding bias metrics in comparison to target metrics in health assessment datasets. \cite{cheng24c_interspeech,gulzar2026biasfairnessselfsupervisedacoustic} identify that MCI detection models can show significant bias across demographic subgroups.


%% file: 030method.tex

\section{Method}\label{sec:prem}
\method consists of two tightly coupled components: 
1) a multimodal MCI classifier with cross-modal fusion to produce accurate MCI diagnosis, and 
2) an unlearning module that removes demographic information from the learned representations to mitigate bias.

\subsection{Multimodal MCI Detection via Cross-Modal Fusion}

Our architecture consists of unimodal encoders followed by a cross-modal (CM) fusion module. Given an input waveform $x_S$, its transcribed text $x_T$, and a corresponding image $x_I$ (if available), we encode them with speech, text, and image encoders $\text{Enc}_S$, $\text{Enc}_T$, and $\text{Enc}_I$, respectively:
\begin{equation}
    z_S = \text{Enc}_S(x_S), \quad z_T = \text{Enc}_T(x_T), \quad z_I = \text{Enc}_I(x_I).
\end{equation}




Prior MCI detection systems often rely on late concatenation~\cite{cheng24c_interspeech,qi2025unveil} to obtain final representations, which underutilizes the rich and fine-grained interactions and dependencies across modalities. To bridge this gap, we employ a cross-attention layer to align and extract interactions between different modalities. Specifically, we perform cross-attention with text as an alignment anchor to fuse (i) linguistic content with paralinguistic cues in speech and (ii) linguistic content with visual semantics when an image is available. For example: 
\begin{equation}
    z = \text{softmax}(\frac{z_S z_T^T}{\sqrt{d_k}} z_T),
\end{equation} where speech features $z_S$ are used as query $Q$, text features $z_T$ are used as key $K$ and values $V$, and $d_k$ is the embedding dimension. The final MCI prediction is given by:
\begin{equation}
    L_{\text{MCI}} = \mathbb{E}_{(x_S, x_T, x_I, y)} \left[ L_{\text{CE}}(y, f_{\text{MCI}}(z)) \right],
\end{equation} where $L_{\text{CE}}$ is the cross-entropy loss.

\textbf{Why CM fusion help?}
This module allows one modality to dynamically attend to the most relevant features of another modality and produce contextually grounded and fine-grained joint embeddings. Specifically, cross-attention allows the model to condition each text token on the most relevant acoustic evidence (e.g., pauses, prosody) and, if available, the most relevant visual cues (e.g., object and scene content). This is meaningful for MCI because predictive signals are often localized and cross-modal: a token may be linguistically correct but acoustically effortful, or visually grounded object naming failures may manifest as vague wording. CM fusion therefore results in a more discriminative and contextually grounded joint representation compared to late concatenation.\looseness-1

\subsection{Unlearning Demographic Biases via Gradient Reversal}\label{sec:unlearn}

MCI detection models are susceptible to demographic biases, i.e., spurious correlations between non-causal demographic information and labels, especially when there is limited training data for MCI tasks. These shortcuts can hurt generalization and amplify performance disparities across patient subgroups and datasets. To mitigate such biases, we explicitly encourage the shared representation $z$ to be predictive of MCI while being {\em uninformative} about demographics.

We introduce an auxiliary demographic classifier $f_{\text{Demo}}$ which predicts demographic attributes from $z$.
To remove reliance on such information, we reverse the gradient when training the demographic classifier, which leads to ``unlearning''\cite{cheng25d_interspeech} the demographic features from the encoder:
\begin{equation}
    L_{\text{Demo}} = \mathbb{E}_{(x,y_d)} \left[ L_{\text{CE}}(y_d, f_{\text{Demo}}(z)) \right],
\end{equation} where $y_d$ is the demographic label provided by the dataset.


We implement the unlearning objective using gradient reversal (GR). During forward propagation, gradient reversal acts as the identity function. During back-propagation, it reverses the gradient with coefficient $\lambda$, effectively reversing the optimization direction for the demographic prediction. This mechanism encourages the shared representation to remove demographic information while remaining predictive for the primary MCI detection task. Intuitively, the encoder learns representations from which the demographic classifier $f_{\text{Demo}}$ cannot reliably infer demographic attributes, thereby discouraging the model from relying on demographic signals to predict MCI.

Formally, the gradient with respect to the shared representation $z$ becomes
\begin{equation}
    \frac{\partial L}{\partial z} = \frac{\partial L_{\text{MCI}}}{\partial z} - \lambda \frac{\partial L_{\text{Demo}}}{\partial z},
\end{equation} where $L_{\text{MCI}}$ is the task loss for MCI prediction and $L_{\text{Demo}}$ is the demographic classification loss.

The demographic adversary can introduce instability during the early stages of training, when task representations are still underdeveloped. We therefore gradually increase the strength of gradient reversal during training for stability~\cite{ganin2015unsupervised}. Specifically, $\lambda$ is scheduled from $0$ to $1$ according to
\begin{equation}
    \lambda = \frac{2}{1+\exp(-\gamma \cdot p)} - 1,
\end{equation} where $\gamma$ controls the rate at which the unlearning strength increases and is treated as a tunable hyperparameter, and $p$ denotes the ratio of current training steps to the total training steps. This is essentially a curriculum~\cite{bengio_cl} that allows the model to first learn task-relevant features before progressively enforcing demographic invariance.

%% file: 040experiments.tex
\section{Experiments}
\subsection{Datasets}
We use the following two datasets for our experiments.
\begin{itemize}
    \item TAUKADIAL \cite{barrera2024interspeech}: a dataset of 387 samples with two labels: normal subjects (NC) and patients with Mild Cognitive Impairment (MCI).
    \item PREPARE \cite{prepare}: a dataset of 1644 samples with three labels: NC, MCI, and ADRD (Alzheimer’s Disease and Related Dementias).
\end{itemize}
The data statistics are shown in Table~\ref{tab:dataset_stats}.

\subsection{Setup}

For TAUKADIAL, we follow \cite{qi2025unveil} to use image, speech, and text (transcribed from the speech files) as inputs. For PREPARE, we use speech and transcribed text as inputs since no image is provided.
Following prior work~\cite{cheng24c_interspeech}, we use 10-fold cross-validation and report the average performance across test folds.
We use Whisper \cite{radford2023robust} as the speech encoder, multilingual BERT \cite{multilingual_bert} as the text encoder, and SigLIP \cite{zhai2023sigmoid} as the image encoder.

\subsection{Evaluation}
The models are trained and evaluated using stratified 10-fold cross-validation.
Following standard bias mitigation literature~\cite{kirichenko2023last,yang2023change,cheng2025students}, we evaluate methods on both the \textbf{overall performance ($\uparrow$)}, \textbf{worst-group performance ($\uparrow$)} of the subgroup and the \textbf{performance gap ($\downarrow$)} between demographic groups. We focus on two demographic attributes: (1) sex (male and female), and (2) language (English and non-English). Ideally, a fair MCI model should have high overall performance, high worst-group performance, and a low performance gap across demographic groups.
For metrics, we use F1 score.


\subsection{Baselines}
We compare \method to the following baselines:
\begin{itemize}
\item Whisper~\cite{radford2023robust}: An encoder-decoder transformer model for speech-to-text tasks. The Whisper encoder is fine-tuned.
\item Audio Spectrogram Transformer (AST)~\cite{gong21b_interspeech}: A vision transformer model for log-mel spectrogram input.
\item XLSR-53~\cite{conneau2021unsupervised}: A multilingual wav2vec 2.0-based speech model pretrained on 53 languages for cross-lingual speech representation learning.
\item XLS-R~\cite{conneau2021unsupervised}: A large multilingual wav2vec 2.0-based speech model trained on massive cross-lingual audio data. We use the 0.3B variant.
\item CogniVoice~\cite{cheng24c_interspeech}: A multilingual and multimodal model that uses a Product of Experts for detecting MCI and predicts MMSE scores from speech and transcripts.
\item DFR~\cite{kirichenko2022last}: A debiasing method that retrains the last layer of a biased model with data from balanced groups.
\item KW~\cite{gao-etal-2022-kernel}: A debiasing method that learns debiased features by disentangling spurious and causal features.
\end{itemize}

%% file: 050results.tex
\begin{table}[t]
\centering
\caption{Dataset statistics.}
\label{tab:dataset_stats}
\footnotesize
\setlength{\tabcolsep}{3.8pt}
\begin{tabular}{l|cccc|cccc}
\toprule
\multirow{3}{*}{\textbf{Label}} 
& \multicolumn{4}{c|}{\textbf{\shortstack{TAUKADIAL \\($n = 387$)}}} 
& \multicolumn{4}{c}{\textbf{\shortstack{PREPARE \\ ($n = 1,644$)}}} \\
\cmidrule{2-9}
& \multicolumn{2}{c}{\textbf{Sex}} & \multicolumn{2}{c|}{\textbf{Language}}
& \multicolumn{2}{c}{\textbf{Sex}} & \multicolumn{2}{c}{\textbf{Language}} \\
\cmidrule(lr){2-3}\cmidrule(lr){4-5}\cmidrule(lr){6-7}\cmidrule(lr){8-9}
& \textbf{F} & \textbf{M} & \textbf{En} & \textbf{Non-En} & \textbf{F} & \textbf{M} & \textbf{En} & \textbf{Non-En} \\
\midrule
NC    & 102 &  63 &  63 & 102 & 539 & 371 &  759 & 151 \\
MCI   & 135 &  87 & 123 &  99 & 123 &  94 &  115 & 102 \\
ADRD  &   - &   - &   - &   - & 297 & 220 &  455 &  62 \\
\midrule
\textbf{Total} & 237 & 150 & 186 & 201 & 959 & 685 & 1,329 & 315 \\
\bottomrule
\end{tabular}
\end{table}



\section{Results}
\subsection{Main Results}

\textbf{\method improves overall performance.}
Across both datasets, \method consistently achieves higher overall F1 scores than all baselines. On TAUKADIAL, \method$^{\textrm{Lang}}$ (where language is used to train $f_{\textrm{Demo}}$) achieves an F1 score of 92.6, substantially outperforming the best baseline, CogniVoice, which achieves an F1 score of 84.1. Similarly, on PREPARE, \method$^{\textrm{Sex}}$ achieves the highest overall F1 score of 60.1, surpassing all baselines. These results demonstrate that \method effectively improves overall MCI detection performance across datasets (see Table~\ref{tab:main_res}).

\textbf{\method mitigates task-irrelevant demographic bias.}
Moreover, \method substantially improves the worst-group F1 (WG), which measures the performance of the most disadvantaged demographic subgroup.
On TAUKADIAL, \method$^{\textrm{Lang}}$ achieves the highest WG F1 of 90.9, significantly higher than the best baseline, CogniVoice, which achieves 81.3 F1. Similarly, on PREPARE, \method$^{\textrm{Lang}}$ achieves the highest WG F1 of 57.4, improving over the best baseline, Whisper, which achieves 54.7 F1. 
%

\method also substantially reduces demographic performance disparities. On TAUKADIAL, \method$^{\textrm{Lang}}$ achieves the lowest average gap of 2.5, outperforming the best-performing baseline, CogniVoice, by 0.4 absolute points. Moreover, \method$^{\textrm{Sex}}$ reduces the sex-group gap to 0.6, compared with the 5.5 sex-group gap of CogniVoice. On PREPARE, \method also achieves the lowest overall demographic gaps, with \method$^{\textrm{Sex}}$ and \method$^{\textrm{Lang}}$ obtaining average gaps of 1.3 and 1.7, respectively, which are both lower than all baselines. In particular, \method$^{\textrm{Lang}}$ achieves the smallest language-group gap of 1.4. These results indicate that \method effectively mitigates task-irrelevant demographic bias while maintaining strong predictive performance. Although CogniVoice achieves a lower gap on the language subgroup than \method on TAUKADIAL, this advantage comes at the expense of significantly lower overall performance (90.9 vs. 81.3 on En, and 92.9 vs. 81.7 on Non-En) and lower worst-group performance (90.9 vs. 81.3)

\begin{table*}[t]
\centering
\vspace{-20pt}
\caption{Performance on the TAUKADIAL and PREPARE datasets measured in F1 score. \method$^{\textrm{w/o UL}}$ denotes the model without the unlearning module. \method$^{\textrm{Sex}}$ and \method$^{\textrm{Lang}}$ denote unlearning applied to the sex and language groups, respectively. Higher F1 ($\uparrow$), higher worst-group F1 (WG, $\uparrow$), and lower gap ($\downarrow$) are better. The best model is in \textbf{bold} and the second best is \underline{underlined}.}
\label{tab:main_res}
\resizebox{\textwidth}{!}{
\begin{tabular}{l|ccc|cccccc|ccc|cccccc}
\toprule
& \multicolumn{9}{c|}{\textbf{TAUKADIAL}} & \multicolumn{9}{c}{\textbf{PREPARE}} \\
\cmidrule(l){2-4} \cmidrule(l){5-7} \cmidrule(l){8-10} \cmidrule(l){11-13} \cmidrule(l){14-16} \cmidrule(l){17-19}
 & \textbf{F1$^\textrm{Avg.}$} & \textbf{WG} & \textbf{Gap$^\textrm{Avg.}$} & \textbf{M} & \textbf{F} & \textbf{Gap} & \textbf{En} & \textbf{Non-En} & \textbf{Gap} & 
 \textbf{F1$^\textrm{Avg.}$} & \textbf{WG} & \textbf{Gap$^\textrm{Avg.}$} & \textbf{M} & \textbf{F} & \textbf{Gap} & \textbf{En} & \textbf{Non-En} & \textbf{Gap} \\
\midrule
Whisper    & 81.3 & 72.2 &  5.3 & 82.0 & 79.9 &  \underline{2.1} & 80.6 & 72.2 &  8.4 & 59.1 & 54.7 & 3.9 & 56.8 & \textbf{60.5} &          3.7 &          58.8 & 54.7 & 4.1 \\
AST        & 71.0 & 53.2 & 16.8 & 77.3 & 66.4 & 10.9 & 76.0 & 53.2 & 22.8 & 58.2 & 51.3 & 4.7 & 58.8 & 57.5 &          1.3 & \textbf{59.5} & 51.3 & 8.2 \\
XLSR-53    & 62.9 & 56.9 & 23.3 & 72.2 & 56.9 & 15.3 & 74.2 & 42.9 & 31.3 & 36.9 & 36.8 & 3.8 & 36.8 & 36.9 & \textbf{0.1} &          38.4 & 45.9 & 7.5 \\
XLS-R      & 75.5 & 61.1 & 12.2 & 78.0 & 72.1 &  5.9 & 79.5 & 61.1 & 18.4 & 38.5 & 35.0 & 2.5 & 38.8 & 38.6 &          \underline{0.2} &          39.7 & 35.0 & 4.7 \\
CogniVoice & 84.1 & 81.3 & \underline{2.9} & 87.8 & 82.3 & 5.5 & 81.3 & 81.7 & \textbf{0.4} & 49.6 & 44.9 & 3.2 & 49.8 & 49.5 & 0.4 & 50.7 & 44.9 & 5.9 \\
DFR        & 83.1 & 81.5 & 3.6 & 85.5 & 81.6 & 3.9 & 81.5 & 84.7 & 3.2 & 53.3 & 50.5 & 4.6 & 55.1 & 51.5 & 3.6 & 56.1 & 50.5 & 5.6 \\
ATG        & 78.6 & 75.2 & 6.8 & 80.3 & 76.9 & 3.4 & 75.2 & 85.4 & 10.2 & 51.5 & 45.1 & 9.1 & 55.0 & 45.1 & 9.9 & 47.4 & 55.6 & 8.2 \\
\midrule
\method$^\textrm{w/o UL}$ & 89.2 & 83.4 & 7.1 & \underline{92.5} & 87.0 & 5.5 & 83.4 & 92.0 & 8.6 & \underline{60.0} & 55.5 & 2.0 & 59.5 & \underline{59.9} & 0.4 & \underline{59.0} & 55.5 & 3.5 \\
\method$^\textrm{Sex}$      & \underline{92.1} & \underline{86.9} & 4.3 & 92.3 & \textbf{91.7} & \textbf{0.6} & \underline{86.9} & \textbf{95.0} & 8.1 & \textbf{60.1} & \underline{56.5} & \textbf{1.3} & \underline{60.1} & 59.8 & 0.3 & 58.5 & \underline{56.5} & \underline{2.0} \\
\method$^\textrm{Lang}$        & \textbf{92.6} & \textbf{90.9} & \textbf{2.5} & \textbf{94.5} & \underline{91.3} & 3.1 & \textbf{90.9} & \underline{92.9} & \underline{2.0} & 59.3 & \textbf{57.4} & \underline{1.7} & \textbf{60.3} & 58.3 & 2.0 & 57.4 & \textbf{58.8} & \textbf{1.4} \\
\bottomrule
\end{tabular}
}
\end{table*}

\begin{table}[t]
\centering
\caption{Comparison of fusion techniques on TAUKADIAL. ``-CM'' and ``-UL'' denote removing the cross-modal (CM) fusion and the unlearning (UL) module, respectively.}
\label{tab:ablation_fusion}
\resizebox{\linewidth}{!}{
\begin{tabular}{l|ccc|ccc|ccc}
\toprule
 & \textbf{F1$^\textrm{Avg.}$} & \textbf{WG} & \textbf{Gap$^\textrm{Avg.}$} & \textbf{M} & \textbf{F} & \textbf{Gap} & \textbf{En} & \textbf{Non-En} & \textbf{Gap} \\
\midrule
\multicolumn{9}{l}{Sex} \\
\midrule
\textbf{\method} & \textbf{92.1} & \textbf{86.9} & \textbf{4.3} & 92.3 & 91.7 & \textbf{0.6} & 86.9 & 95.0 & \textbf{8.1} \\
 - CM   & 90.7 & 83.4 & 8.3 & 94.2 & 88.3 & 5.9 & 83.4 & 94.0 & 10.6 \\
 - UL   & 89.2 & 83.4 & 7.1 & 92.5 & 87.0 & 5.5 & 83.4 & 92.0 & 8.6 \\
\midrule
\multicolumn{9}{l}{Language} \\
\midrule
\textbf{\method} & \textbf{92.6} & \textbf{90.9} & \textbf{2.5} & 94.5 & 91.3 & \textbf{3.1} & 90.9 & 92.9 & \textbf{2.0} \\
 - CM   & 91.5 & 86.0 & 7.7 & 95.6 & 88.7 & 6.9 & 86.0 & 94.5 & 8.5 \\
 - UL   & 89.2 & 83.4 & 7.1 & 92.5 & 87.0 & 5.5 & 83.4 & 92.0 & 8.6 \\
\bottomrule
\end{tabular}
}
\end{table}

\begin{table}[t]
\centering
\caption{Transfer performance. A $\rightarrow$ B indicates training on dataset A and testing on dataset B. CogniVoice is the best-performing baseline.}
\label{tab:transfer}
\resizebox{\linewidth}{!}{
\begin{tabular}{l|ccc|ccc|ccc}
\toprule
 & \textbf{F1$^\textrm{Avg.}$} & \textbf{WG} & \textbf{Gap$^\textrm{Avg.}$} & \textbf{M} & \textbf{F} & \textbf{Gap} & \textbf{En} & \textbf{Non-En} & \textbf{Gap} \\
\midrule
\multicolumn{9}{l}{TAUKADIAL $\rightarrow$ PREPARE} \\
\midrule
CogniVoice                & 38.7 & 35.5 & 4.1 & 39.6 & 37.8 & 1.8 & 41.9 & 35.5 & 6.4\\
\method$^\textrm{w/o UL}$ & 41.2 & \underline{39.8} & \textbf{2.7} & 40.1 & 41.2 & 1.1 & 39.8 & 45.2 & 5.4 \\
\method$^\textrm{Sex}$    & \textbf{42.3} & \textbf{41.0} & \underline{3.3} & 41.0 & 42.9 & 1.9 & 41.2 & 45.6 & 4.4 \\
\method$^\textrm{Lang}$   & \underline{41.3} & 39.5 & 4.4 & 40.6 & 41.6 & 1.0 & 39.5 & 47.2 & 7.7 \\
\midrule
\multicolumn{9}{l}{PREPARE $\rightarrow$ TAUKADIAL} \\
\midrule
CogniVoice                & 39.8 & 26.8 & 19.1 & 33.7 & 45.9 & 12.2 & 26.8 & 52.8 & 26.0 \\
\method$^\textrm{w/o UL}$ & 44.8 & \textbf{33.0} & \underline{16.8} & 38.4 & 48.6 & 10.2 & 56.3 & 33.0 & 23.3 \\
\method$^\textrm{Sex}$    & \underline{45.2} & \underline{32.9} & \textbf{16.7} & 39.8 & 48.3 &  8.5 & 57.8 & 32.9 & 24.9 \\
\method$^\textrm{Lang}$   & \textbf{45.5} & \textbf{33.0} & \textbf{16.7} & 40.4 & 48.5 &  8.1 & 58.3 & 33.0 & 25.3 \\
\bottomrule
\end{tabular}
\vspace{-10pt}
}
\end{table}

\subsection{Ablation Study}
We conduct a two-fold ablation study on the modality fusion module and the unlearning module. As Table~\ref{tab:ablation_fusion} shows, removing the cross-modal (CM) fusion module leads to performance degradation. Specifically, the overall F1 score drops by 1.4 and 1.1 absolute points in the sex and language groups, respectively. Additionally, the subgroup gap becomes more prominent after removing CM, with increases of 4.0 and 5.2 absolute points in the sex and language groups, respectively. We also observe a worse worst-group F1 score when ablating CM. These results highlight the effectiveness of CM, indicating that it provides finer-grained and better-aligned information across modalities.

In addition, removing the unlearning (UL) module leads to noticeable performance degradation. Specifically, for the sex subgroup, the overall F1 score and the average gap decrease by 2.9 and 2.8 absolute points, respectively. For the language subgroup, the overall F1 score and the average gap decrease by 2.4 and 4.6 absolute points, respectively. These results highlight the effectiveness of the unlearning module in mitigating task-irrelevant demographic bias while encouraging the model to focus on more indicative signals.

\subsection{Transfer Performance}
To further examine the robustness of the models, we conduct a zero-shot transfer experiment, where the model is trained on one dataset and evaluated on another.
When training on TAUKADIAL and testing on PREPARE, all variants of \method achieves higher overall F1 scores than the best-performing baseline CogniVoice. In particular, \method$^{\textrm{Sex}}$ outperforms CogniVoice by 3.6 and 0.8 absolute point in overall F1 and F1 gap respectively, demonstrating improved generalization under distribution shift. We observe similar trends when training on PREPARE and testing on TAUKADIAL. All variants of \method outperform CogniVoice, with higher overall performance and reduced performance gap.

These results suggest that \method encourages the model to rely on more task-relevant signals to detect MCI rather than dataset-specific biases, leading to better generalization.\looseness-1


\begin{table}[t]
\centering
\caption{Bias probing.
A performance of 50\% indicates that the representation contains less task-irrelevant demographic information.}
\label{tab:probing}
\vspace{-5pt}
\footnotesize
\begin{tabular}{l|cc}
\toprule
\textbf{Method} & \textbf{Sex ($\downarrow$)} & \textbf{Lang ($\downarrow$)} \\
\midrule
CogniVoice & 71.2 & 68.5 \\
\method    & 61.7 & 62.3 \\
\bottomrule
\end{tabular}
\vspace{-15pt}
\end{table}

\subsection{Demographic Bias Probing}
To further evaluate whether the unlearning module can remove task-irrelevant demographic information, we train additional probing classifiers (logistic regression) to predict demographic labels from the model representations $z$ for each model. An accuracy closer to random guessing (50\%) indicates that the representations contain less demographic bias.
As shown in Table~\ref{tab:probing}, \method reduces demographic information leakage compared to the best-performing baseline, CogniVoice. For the sex subgroup, the probing classifier achieves an accuracy of 71.2 on the representations of CogniVoice, while it achieves 61.7 on \method, which is closer to random guessing. Similarly, for the language subgroup, the probing accuracy is 68.5 for CogniVoice and 62.3 for \method. These reductions suggest that the unlearning mechanism in \method suppresses demographic predictive signals in the shared representations and encourages greater reliance on task-relevant signals; however, the probe accuracies remain above random guessing, suggesting that residual demographic information is still present.


%% file: 060conclusion.tex
\section{Conclusion}
We study demographic bias in Mild Cognitive Impairment (MCI) detection, where models rely on spurious demographic attributes and show substantial performance disparities across different patient subgroups.
We propose \method, a framework that combines cross-modal representation fusion and an unlearning designed to discourage demographic information from being encoded in the shared representation in MCI detection. Across two benchmarks, \method achieves improved overall predictive performance while reducing performance disparities across demographic subgroups. Furthermore, \method learns representations that transfer more effectively across datasets, suggesting more robustness and a stronger emphasis on disease-relevant features compared to baselines. While we show residual leakage remains, these results support \method as a practical step toward fair and more generalizable MCI screening models.





\section{Generative AI Use Disclosure}

We used LLMs to assist with editing and polishing our writing for clarity and concision. All technical content, analyses, and conclusions were produced by the authors and verified for accuracy.\looseness-1